\documentclass{article}

\usepackage{arxiv}
\usepackage{natbib}
\usepackage[utf8]{inputenc} 
\usepackage[T1]{fontenc}    
\usepackage{hyperref}       
\usepackage{url}            
\usepackage{booktabs}       
\usepackage{amsfonts}       
\usepackage{nicefrac}       
\usepackage{microtype}      
\usepackage{lipsum}
\usepackage{graphicx}
\graphicspath{ {./images/} }

\usepackage{hyperref}
\usepackage{url}
\usepackage{bm}
\usepackage{booktabs}
\usepackage{multirow}
\usepackage{graphicx}
\usepackage{subfigure}
\usepackage{amsmath}
\usepackage{enumitem}

\def \x {\bm{x}}

\title{What Makes ``Good'' Distractors for Object Hallucination Evaluation in Large Vision-Language Models?}

\author{
  Ming-Kun Xie \\
  Center for Advanced Intelligence Project, RIKEN \\
  \texttt{ming-kun.xie@riken.jp} \\
  \And
  Jia-Hao Xiao \\
  Southeast University \\
  \texttt{jiahaoxiaooo@gmail.com} \\
  \And
  Gang Niu \\
  Center for Advanced Intelligence Project, RIKEN \\
  \texttt{gang.niu.ml@gmail.com} \\
  \And
  Lei Feng \\
  Southeast University \\
  \texttt{fenglei@seu.edu.cn} \\
  \And
  Zhiqiang Kou \\
  Southeast University \\
  \texttt{zhiqiang\_kou@seu.edu.cn} \\
  \And
  Min-Ling Zhang \\
  Southeast University \\
  \texttt{zhangml@seu.edu.cn} \\
  \And
  Masashi Sugiyama \\
  Center for Advanced Intelligence Project, RIKEN \\
  The University of Tokyo\\
  \texttt{sugi@k.u-tokyo.ac.jp} \\
}

\begin{document}
\maketitle
\begin{abstract}
Large Vision-Language Models (LVLMs), empowered by the success of Large Language Models (LLMs), have achieved impressive performance across domains.
Despite the great advances in LVLMs, they still suffer from the unavailable object hallucination issue, which tends to generate objects inconsistent with the image content. The most commonly used Polling-based Object Probing Evaluation (POPE) benchmark evaluates this issue by sampling negative categories according to category-level statistics, \textit{e.g.}, category frequencies and co-occurrence. However, with the continuous advancement of LVLMs, the POPE benchmark has shown diminishing effectiveness in assessing object hallucination, as it employs a simplistic sampling strategy that overlooks image-specific information and restricts distractors to negative object categories only. In this paper, we introduce the Hallucination searching-based Object Probing Evaluation (HOPE) benchmark, aiming to generate the most misleading distractors (\textit{i.e.}, non-existent objects or incorrect image descriptions) that can trigger hallucination in LVLMs, which serves as a means to more rigorously assess their immunity to hallucination. To explore the image-specific information, the content-aware hallucination searching leverages Contrastive Language-Image Pre-Training (CLIP) to approximate the predictive behavior of LVLMs by selecting negative objects with the highest predicted likelihood as distractors. 
To expand the scope of hallucination assessment, the description-based hallucination searching constructs highly misleading distractors by pairing true objects with false descriptions. Experimental results show that HOPE leads to a precision drop of at least 9\% and up to 23\% across various state-of-the-art LVLMs, significantly outperforming POPE in exposing hallucination vulnerabilities. The code is available at \url{https://github.com/xiemk/HOPE}.
\end{abstract}


\section{Introduction}

Large Language Models (LLMs), exemplified by chatbots such as ChatGPT \citep{gpt4o}, Gemini \cite{team2023gemini}, and DeepSeek \citep{liu2024deepseek}, have undergone rapid advancements and now exhibit a remarkable ability to comprehend and respond to human queries. The success of LLMs in the textual modality has inspired the research community to explore foundation models capable of integrating multiple modalities, thereby significantly accelerating the progress in Large Vision-Language Models (LVLMs). The success of LVLMs has been demonstrated across a wide range of practical applications, including but not limited to  biomedical visual question answering \citep{li2023llava} and embodied AI \citep{driess2023palm}.

Despite the remarkable progress achieved by LVLMs, they still suffer from a critical and unavoidable issue known as object hallucination \citep{biten2022let,li2023evaluating}, \textit{i.e.}, the tendency to generate objects inconsistent with the image content. This issue is particularly critical because it can significantly undermine user trust in chatbot systems \citep{ji2023survey}, which in turn affects whether users are willing to engage with them regularly. Therefore, before deploying an LVLM-based product, it is essential to assess the extent to which the model exhibits object hallucination. 

The most widely adopted benchmark for evaluating object hallucination is Polling-based Object Probing Evaluation (POPE) \citep{li2023evaluating}. It assesses object hallucination by sampling a fixed number of positive and negative object categories from each image, and prompting LVLMs with binary questions of the form ``Is there a \{object\} in the image?'' regarding the presence of these objects. 
Although this benchmark has been widely used for evaluating object hallucination, our study reveals that as LVLMs continue to evolve through successive iterations, the hallucination problem has been significantly mitigated. As a result, the POPE benchmark is becoming increasingly outdated and fails to precisely reflect the hallucination behavior of state-of-the-art LVLMs. We identify two main limitations of POPE: (i) Simplistic sampling strategy: POPE relies solely on category-level statistics, \textit{e.g.}, object frequency or co-occurrence, when sampling negative object categories. It does not account for image-specific content, and thus fails to construct hallucination targets that are tailored to each individual image. (ii) Narrow sampling space: POPE only considers negative objects (\textit{i.e.}, categories not present in the image) as hallucination candidates, ignoring other plausible sources of hallucination. 
Due to these limitations, the distractors generated by POPE are often insufficiently misleading, making it difficult to accurately assess the hallucination behavior of state-of-the-art LVLMs. 
In response to the second limitation, existing efforts such as H-POPE \citep{pham2024h} and R-Bench \citep{wu2024evaluating} have extended POPE to evaluate hallucinations at the attribute and relationship levels, respectively. However, these benchmarks do not consider how to construct the most misleading distractors in an instance-dependent manner, which is a central aspect that our work explicitly addresses.
In practice, models may appear robust not because they have overcome the hallucination problem, but because the evaluation benchmark fails to challenge them effectively. 
Therefore, it is crucial to construct a more challenging and up-to-date benchmark to rigorously evaluate object hallucination in modern LVLMs.



To address this challenge, we formalize the evaluation process as an optimization problem, whose goal is to search ``good distractors'' that are most likely to induce hallucination. The underlying insight is to treat object hallucination evaluation as a contest between spear and shield: only the sharpest spears (distractors) can truly test the strength of shields (LVLMs). That is, the immunity of an LVLM to hallucination can only be meaningfully assessed when it is exposed to highly misleading and carefully designed distractors.
Given that the target LVLMs are inaccessible, we propose a top-$k$ heuristic strategy to approximate the original optimization process. It leverages a hallucination scorer to estimate the likelihood of hallucination, enabling us to efficiently search the most misleading distractors without querying the target LVLM directly. We refer to this process as \textit{hallucination searching}. 

In this framework, the effectiveness of the searching strategy primarily depends on the definition of the hallucination scorer, which entails two essential aspects: the design of the distractor space and the computation of the hallucination probability for each candidate distractor. Specifically, we develop three hallucination searching strategies: \textit{category-oriented}, \textit{content-aware}, and \textit{description-based}. 
Category-oriented searching serves as a basic strategy, relying primarily on relationships between object categories. Besides object co-occurrence, which has already been employed in POPE, we introduce a complementary visual similarity metric to identify the most similar negative object as a distractor for each positive object. 
Since prior strategies are instance-independent and thus unable to generate distractors uniquely suited to each image, our main contribution is the introduction of two instance-dependent search strategies that leverage image-specific information to better reveal hallucination vulnerabilities.
On one hand, to further enhance the effectiveness of searching strategies, we develop the content-aware searching, which introduces Contrastive Language-Image Pre-Training (CLIP) as a proxy to approximate the prediction behavior of LVLMs. By incorporating image-specific information, this strategy identifies the most deceptive negative categories based on ambiguous visual content present in the image. 
On the other hand, to further expand the distractor space beyond negative object categories, we propose the description-based searching, which generates more challenging distractors by pairing true objects with false descriptions (\textit{e.g.}, incorrect attributes or states). These distractors are further refined based on their alignment with the image content, enhancing their ability to mislead LVLMs. 
Experimental results with multiple state-of-the-art LVLMs verify the effectiveness of the proposed method.

\section{Problem Formulation}





In this section, we present a formalization of hallucination searching, with the goal of generating distractors that are most likely to induce hallucination in LVLMs. Let $\x\in\mathcal{X}$ represents an input image, where $\mathcal{X}$ is the feature space, and $P\subset{\mathcal{Y}}$ and $N\subset{\mathcal{Y}}$ are its positive and negative object category sets, where $\mathcal{Y}$ is the category space.  A prompt template is denoted as $\tau\in\mathcal{T}$, where $\mathcal{T}$ is the set of all available prompt templates. 
The distractor space $\mathcal{D}$ may include not only negative object categories $N$, \textit{e.g.}, ``car'', but also object-description pairs, where descriptions are drawn from the predefined set $\mathcal{A}$, such as attributes (\textit{e.g.}, ``red sport car'') or states (\textit{e.g.}, ``car running on the road''). Formally, we define the optimization objective as
\begin{equation}
    \max_{D\subset{\mathcal{D}},|D|\leq k}\mathcal{L}_{\text{Hal}}(D; \x,P,\tau),
\end{equation}
where $\mathcal{L}_{\text{Hal}}$ is a measurement that quantifies how many elements in $D$ are hallucinated by the LVLM, given the input image $\x$, the positive object categories $P$,  and the prompt template $\tau$. Considering the cost associated with LVLM evaluation, such as token-based billing in closed-source models or substantial computation and time in open-source settings, the parameter $k$ serves as an upper bound on the size of the distractor set. This constraint enables cost-efficient yet effective hallucination evaluation. However, directly optimizing $\mathcal{L}_{\text{Hal}}$ is infeasible in practice, as it depends on the inaccessible behavior of the target LVLMs, which may vary across users and model versions.

To address this issue, we introduce a hallucination scorer $h(d; \x, P)$ that estimates the probability that a given distractor $d \in \mathcal{D}$ will be hallucinated. Using this scorer, we construct the distracotr set $D$ via a top-$k$ selection:
\begin{equation}
    \mathcal{L}_{\text{Hal}}(D; \x,P,\tau),\quad \text{where}\quad D={\text{Top-}k}_{d\in \mathcal{D}}(h(d;\x,P)).
\end{equation}
That is, we select $k$ distractors with the highest hallucination scores. Note that a distractor $d$ is not necessarily related to the input image $\x$ or the set of positive categories $P$. The definition of the hallucination scorer $h$ depends on the specific searching strategy, which we will discuss in detail in the following sections. Generally, this heuristic allows us to approximate the ideal distractor set $D$ without relying on access to the target LVLM, enabling scalable and model-agnostic evaluation. 
From the formalization, we can see that performance of searching strategies on LVLMs depends on three components: the hallucination scorer $h$, the size of the distractor space $\mathcal{D}$, and the prompt template $\tau$. Generally, the hallucination scorer 
$h$ determines how precisely we can estimate the probability that each candidate disctractor will induce hallucination in LVLMs, and the size of the distractor space affects whether it is likely to contain more ``good'' distractors that are effective in triggering hallucination.


\section{HOPE Benchmark}


We first introduce the Hallucination searching-based Object Probing Evaluation (HOPE) benchmark, and then discuss the prompt templates used in this benchmark.

\subsection{Hallucination Searching}

The goal of hallucination searching is to identify distractors that are most likely to induce hallucination in LVLMs. Specifically, we design three searching strategies: Category-oriented hallucination searching, which focuses on identifying negative categories that are likely to induce hallucination, conditioned on the objects present in the image; content-aware hallucination searching, which targets ambiguous visual content within the image that may lead to incorrect object predictions; description-based hallucination searching, which targets spurious object-description combinations to trigger hallucination. Below, we introduce each of these hallucination searching strategies in detail.


\subsubsection{Category-Oriented Hallucination Searching}

In this strategy, we perform hallucination searching targeting each ground-truth object $p\in P$ present in the image, aiming to identify a distractor $d\in N$, \textit{i.e.}, a negative category, that is most likely to induce hallucination in LVLMs. For each image, the distractor candidate set $D$ reduces to the negative category set $N$. 

To quantify the hallucination potential of each distractor, we define a pairwise hallucination scorer $h(d,p)$ to estimate the probability that each negative category $d\in N$ will induce hallucination, given the presence of the ground-truth category $p$. Considering that an LVLM typically consists of an LLM and an image encoder, we instantiate the hallucination scorer $h$ based on two complementary cues: object co-occurrence and visual similarity. The former reflects language prior knowledge, namely, objects that frequently co-occur in the physical world are also likely to appear together in natural language descriptions. The similar idea has been used in the POPE benchmark \citep{li2023evaluating}. The latter captures visual prior knowledge, that is, objects with similar appearances are more likely to be confused by the model. To obtain visual similarity, we adopt a simple yet effective approach by computing the similarity between the category names encoded by the CLIP text encoder. This is motivated by the fact that CLIP aligns image and text representations in a shared embedding space, so the similarity between textual features can serve as a reasonable proxy for visual similarity. 

Formally, we define the co-occurrence-based and similarity-based pairwise hallucination scorers:
\begin{equation*}
    \forall d\in N, p\in P, h_\text{coo}(d,p)=\frac{\text{Count}(p,d)}{\text{Count}(p)}, h_\text{sim}(d,p)=\text{Cosine}(E_\text{T}(\tau_{\text{CLIP}}(d)),E_\text{T}(\tau_{\text{CLIP}}(p))),
\end{equation*}
where $\text{Count}(p,d)$ and $\text{Count}(p)$ respectively denote the number of images that contain both objects $p$ and $d$ and the number of images that contain object $p$ alone. $E_\text{T}(\cdot)$ represents the text encoder in CLIP, $\tau_{\text{CLIP}}(\cdot)$ is the template used in CLIP, and $\text{Cosine}(\cdot,\cdot)$ computes the cosine similarity.




\subsubsection{Content-Aware Hallucination Searching}

While category-oriented hallucination searching focus on relationships between object categories, the searching strategy for a specific object remains consistent across images. However, images may vary significantly in content, and beyond the explicit objects present, there may exist ambiguous visual patterns or regions that can also serve as effective targets for hallucination. Therefore, searching based directly on image contents represents a natural and complementary strategy. 

To ensure that the benchmark is applicable across a broad range of LVLMs, we adopt CLIP as a surrogate model for hallucination scoring. This choice is motivated by the fact that CLIP, like most LVLMs, is trained on large-scale image-text pairs, and consequently exhibits similar prediction behavior. Therefore, it serves as a reasonable and efficient proxy for estimating the likelihood that a negative object category will be hallucinated, without requiring direct access to the target LVLM. \

Specifically, given an image $\x$, our objective is to identify the most suspicious negative category $d\in N$, based on the ambiguous content present in the image. Formally, we define the content-aware hallucination scorer as
\begin{equation*}
    \forall d\in N, h_{\text{con}}(\x, d)=\text{Cosine}(E_{\mathrm{I}}(\x),E_{\mathrm{T}}(\tau_{\text{CLIP}}(d))),
\end{equation*}
where $E_{\mathrm{I}}(\cdot)$ and $E_{\mathrm{T}}(\cdot)$ represent the image encoder and text encoder in CLIP.

\begin{figure*}[!t]
    \centering
    {\includegraphics[width=0.99\textwidth]{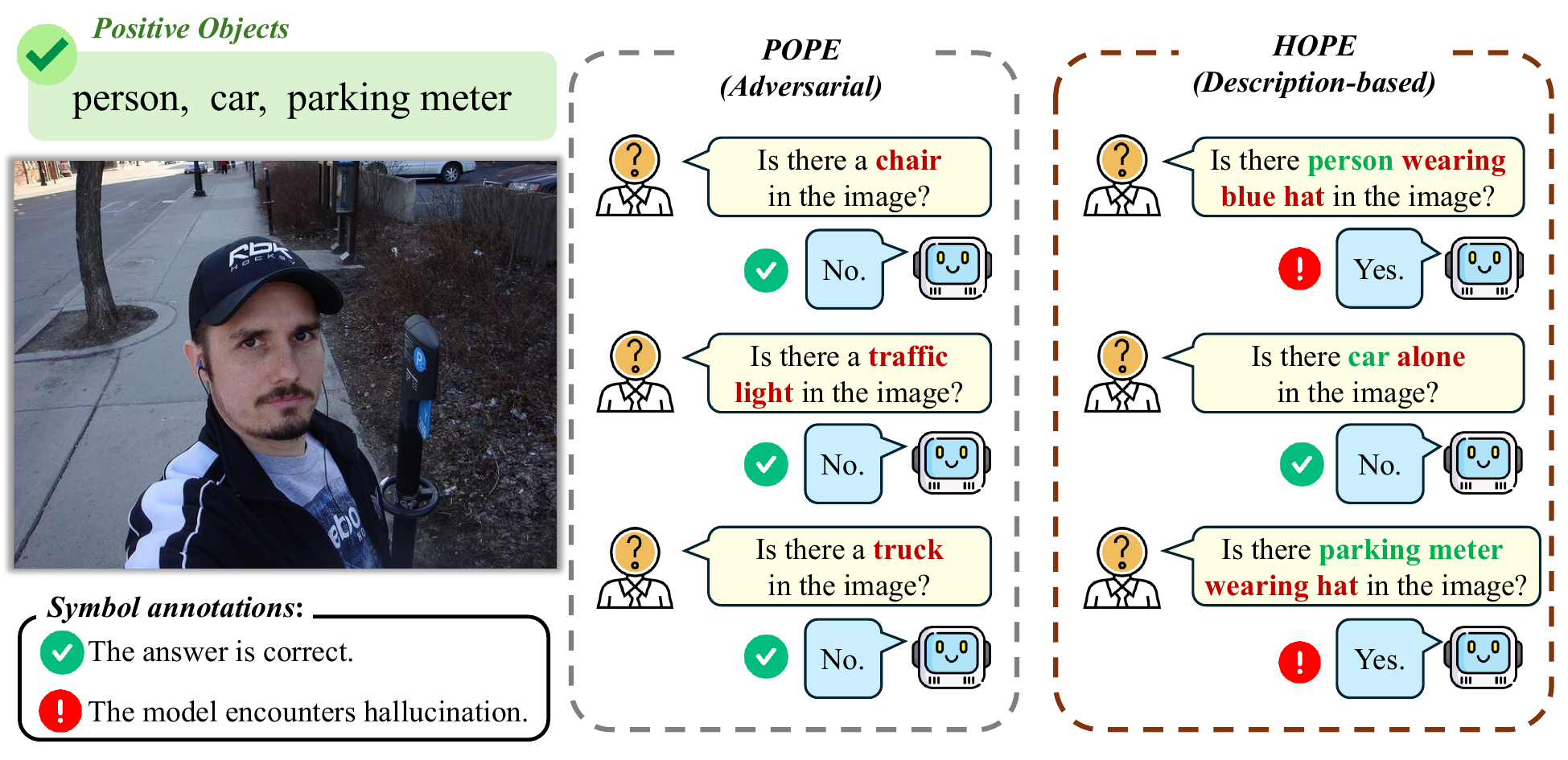}}
    \caption{An illustrative example of comparision between POPE and HOPE. }
    \label{fig:attribute}
\end{figure*}


\subsubsection{Description-Based Hallucination Searching}

The two searching strategies discussed above both aim to identify the most misleading negative object categories.  These strategies can be viewed as category-level hallucination searching, as they require LVLMs to precisely distinguish between different semantic concepts. 

Here, we introduce a more challenging setting for LVLMs: the description-based hallucination searching. An question that we explore is whether it is possible to construct distractors that retain true objects while still inducing hallucination. Intuitively, such distractors are particularly deceptive because they contain the true object but may mislead the LVLM due to spurious descriptions.

To construct description-based distractors, for an given image $\x$ and a positive object category $p$, we assume that there is a set of negative description candidates 
$\mathcal{A}(\x,p)$, which consists of descriptions that are related to object $p$ but semantically irrelevant to the visual content of the image. In practice, we construct the negative description candidate set $\mathcal{A}(\x, p)$ by collecting all descriptions associated with object $p$ across the dataset, and removing those that appear in the target image. After obtaining the negative description set $\mathcal{A}(\x,p)$, our goal is to identify the object-description combinations that are most likely to induce hallucination in the LVLM. 
Formally, we define the description-based hallucination scorer as
\begin{equation*}
    h_{\text{attr}}(\x, d)=\text{Cosine}(E_{\mathrm{I}}(\x),E_{\mathrm{T}}(\tau_{\text{CLIP}}(d))),\quad \text{where} \quad d=\text{Concat}(p,a), \forall a\in\mathcal{A}(\x,p), p\in P,
\end{equation*}
and $\text{Concate}(\cdot,\cdot)$ denotes an operator to form object-description phrases by placing the attribute either before or after the object. Specifically, adjectives such as colors are placed before the object (\textit{e.g.}, “red car”), while verbs or action-related attributes are placed after it (\textit{e.g.}, “car running”). Given that similar descriptions may appear across different images, we additionally perform manual verification to ensure that each constructed distractor $d$ contains only descriptions that are semantically inconsistent with the corresponding image.


\subsection{Prompt Template}

Given the flexibility of natural language, the number of possible prompt templates is virtually unlimited. In our study, we categorize prompt templates into two primary types: \textit{binary} prompts and \textit{multi-option} prompts. The binary prompt template has been adopted in the POPE benchmark. It typically takes the form of asking an LVLM whether a specific object is present in the image, such as ``Is there a \{object\} in the image?'' The LVLM is only required to respond with a binary answer: ``Yes'' or ``No''. In contrast, the multi-option prompt template presents the LVLM with a predefined set of candidate objects and asks which ones are present in the image, \textit{e.g.}, ``What objects are present in the image? The candidate set is: $\{\text{object}_1\}$, \ldots, $\{\text{object}_k\}$”, where $k$ is the number of candidate objects. The LVLM is required to provide the specific object categories it identifies as present in the image. 

Compared to binary prompts, processing multi-option prompts is evidently more challenging for LVLMs. This increased difficulty can be attributed to two main factors:
(i) Prompt complexity: Multi-option prompts are longer and more information-dense, as they explicitly include a list of candidate object categories.
(ii) Output space expansion: For binary prompts, the model only needs to choose between two possible responses (``Yes'' or ``No''), whereas multi-option prompts require selecting one or more categories from the candidate set, resulting in a combinatorially larger output space. On the other hand, providing candidate object categories as contextual information can be either beneficial or detrimental to LVLM performance. The actual effect depends on both the specific LVLM being used and the strategy employed to perform hallucination searching. We extend the POPE framework by incorporating multi-option prompts, and conduct experiments to perform an in-depth analysis of LVLM performance under this prompt template.

\section{Experiment}

In this section, we first introduce the detailed experimental setup, followed by a presentation of the results, including both main comparative evaluations and qualitative visualizations.

\subsection{Experimental Settings}

To construct the HOPE benchmark, we use the Objects365 \footnote{\url{https://www.objects365.org/overview.html}} \citep{shao2019objects365} dataset for the category-oriented and content-aware hallucination searching, as it includes all object categories present in MS-COCO and contains 365 categories in total, which is significantly more than the 80 categories available in MS-COCO \citep{lin2014microsoft}. The dataset is originally divided into training and validation sets, and and we directly use its validation sets in our experiments. We randomly sample 2,000 images and select 6 positive labels per image, as the average number of object categories per image in the dataset is about 6. For each searching strategy, including the co-occurrence and similarity variants of the category-oriented searching as well as the content-aware searching, we select 6 negative object categories with the highest hallucination scores. Moreover, we construct a hardest setting by sequentially applying all three hallucination searching. For each image, we iteratively select the top-scoring negative object categories based on hallucination scores from each searching strategy, in the order of co-occurrence, similarity, and content-aware. We add only those distractors that have not yet being included in the current set. For the description-based hallucination searching, we use Large-Scale Attribute (LSA) \citep{Pham} dataset, as it provides fine-grained annotations such as attributes, states, and relationships. The LSA dataset is constructed by extracting object-centric attributes and interactions from three existing datasets, MS-COCO 2014 \footnote{\url{https://cocodataset.org/\#home}} \citep{lin2014microsoft}, Visual Genome (VG) \footnote{\url{https://homes.cs.washington.edu/~ranjay/visualgenome/index.html}} \citep{krishna2017visual}, and OpenImages \footnote{\url{https://storage.googleapis.com/openimages/web/index.html}} \citep{kuznetsova2020open}, all of which contain paired image-text data. For each image in the dataset, we randomly select three ground-truth object categories and identify the top three distractors with the highest hallucination scores. Each distractor corresponds to a true-object and false-attribute combination. Following the previous work \citep{li2023evaluating}, we use three evaluation metrics: the precision, recall, and F1 score.

We evaluate our benchmark using three representative families of state-of-the-art LVLMs: Qwen-VL series,
including Qwen2-VL \citep{wang2024qwen2} and Qwen2.5-VL \citep{bai2025qwen2}, LLaVA-NeXT series \citep{liu2024llavanext},
including LLaVA-Next and LLaVA-OneVision (OV), InternVL series,
including InternVL2.5 \citep{chen2024expanding} and InternVL3 \citep{zhu2025internvl3}.

\begin{table}[!t]
  \centering
  \small
  \caption{Comparative Results with the state-of-the-art LVLMs between POPE and HOPE on Objects365. \textit{Binary} and \textit{Multi-Option} denote two types of prompt templates.}
    \begin{tabular}{c|l|ccc|ccc}
    \toprule
    \multicolumn{1}{c}{\multirow{2}[4]{*}{Strategy}} & \multicolumn{1}{c|}{Prompt} & \multicolumn{3}{c|}{ Binary} & \multicolumn{3}{c}{Multi-Option} \\
    \cmidrule{2-8}    \multicolumn{1}{c}{} & \multicolumn{1}{c|}{Model} & Precision & Recall & F1 Score & Precision & Recall & F1 Score \\
    \midrule
    \multirow{6}[2]{*}{\shortstack{Co-occurrence\\(\textit{Adversarial} in POPE)} } & LLaVA Next 8B & 81.94  & 77.02  & 79.40  & 76.14  & 71.94  & 73.98  \\
          & LLaVA OV 7B & 81.16  & 79.85  & 80.50  & 87.69  & 58.28  & 70.02  \\
          & Qwen2-VL 7B & 85.85  & 74.39  & 79.71  & 79.04  & 74.97  & 76.95  \\
          & Qwen2.5-VL 7B & 86.78  & 70.39  & 77.73  & 71.86  & 80.35  & 75.87  \\
          & InternVL2.5 8B & 84.45  & 76.14  & 80.08  & 80.74  & 74.98  & 77.75  \\
          & InternVL3 8B & 83.71  & 80.59  & 82.12  & 83.22  & 78.33  & 80.70  \\
    \midrule
    \multirow{6}[2]{*}{Similairity} & LLaVA Next 8B & 76.47  & 77.02  & 76.74  & 75.12  & 71.97  & 73.51  \\
          & LLaVA OV 7B & 77.71  & 79.85  & 78.76  & 88.46  & 57.13  & 69.42  \\
          & Qwen2-VL 7B & 80.89  & 74.39  & 77.50  & 77.97  & 74.13  & 76.00  \\
          & Qwen2.5-VL 7B & 85.31  & 70.39  & 77.13  & 75.81  & 79.93  & 77.81  \\
          & InternVL2.5 8B & 81.77  & 76.14  & 78.86  & 78.45  & 81.13  & 77.48  \\
          & InternVL3 8B & 80.12  & 80.59  & 80.36  & 84.15  & 78.04  & 80.98  \\
    \midrule
    \multirow{6}[2]{*}{Content-Aware} & LLaVA Next 8B & 71.49  & 77.02  & 74.15  & 67.63  & 66.07  & 66.84  \\
          & LLaVA OV 7B & 72.36  & 79.85  & 75.92  & 80.90  & 55.87  & 66.09  \\
          & Qwen2-VL 7B & 76.51  & 74.39  & 75.44  & 72.01  & 71.91  & 71.96  \\
          & Qwen2.5-VL 7B & 81.39  & 70.39  & 75.49  & 69.79  & 79.27  & 74.23  \\
          & InternVL2.5 8B & 77.72  & 76.14  & 76.92  & 74.51  & 71.55  & 73.00  \\
          & InternVL3 8B & 77.00  & 80.59  & 78.76  & 77.71  & 75.43  & 76.55  \\
          \midrule
          \multirow{6}[6]{*}{All} & LLaVA Next 8B & 54.53  & 77.02  & 63.85  & 49.78  & 73.42  & 59.33  \\
          & LLaVA OV 7B & 55.12  & 79.85  & 65.33  & 62.21 & 60.00  & 61.08  \\
          & Qwen2-VL 7B & 61.52  & 74.39  & 67.35  & 49.03 & 74.97  & 59.29  \\
          & Qwen2.5-VL 7B & 67.83  & 70.39  & 69.05  & 47.58  & 79.88  & 59.63  \\
          & InternVL2.5 8B & 62.81  & 76.14  & 68.84  & 60.19  & 75.34  & 66.92  \\
          & InternVL3 8B & 60.86  & 80.59  & 69.35  & 60.79  & 80.96  & 69.44  \\
    \bottomrule
    \end{tabular}%
  \label{tab:o365}%
\end{table}%

\subsection{Results of State-of-The-Art LVLMs}

Table \ref{tab:o365} reports the results of different LVLMs under three hallucination searching srategies on the Objects365 dataset using the $\textit{Binary}$ and $\textit{Multi-Option}$ prompt templates. Note that the sampled positive object categories are consistent across all searching strategies. Under the binary prompt setting, changing the negative categories does not affect the model's predictions on positives. Therefore, the recall remains identical for each LVLM across all three searching strategies. From the table, we observe that:
(i) The success rate of the three searching strategies increases progressively, as evidenced by the gradual decline in model precision. The content-aware searching achieves the highest success rate, indicating that searching relying solely on category-level information without considering visual features—as done in POPE—are less effective. In contrast, targeting ambiguous contents within the image is more likely to induce hallucination in LVLMs.
(ii) Compared to any individual searching, the combination of all three searching strategies leads to a significant drop in model precision. This suggests that the distractors generated by the different strategies are largely complementary and have minimal overlap. By combining these strategies, we construct a highly challenging evaluation setting for object hallucination evaluation. 
(iii) Compared to the binary prompt, the multi-option prompt is clearly more challenging and more likely to induce hallucination in LVLMs. As discussed earlier, on one hand, the output space of the LVLM is significantly expanded; on the other hand, the prompt introduces contextual information from the candidate categories, which often misleads the model.

\begin{table}[!t]
  \centering
  \caption{Comparative results with the state-of-the-art LVLMs between HOPE and POPE on MS-COCO.}
    \begin{tabular}{l|ccc|ccc}
    \toprule
    \multicolumn{1}{c|}{Benchmark} & \multicolumn{3}{c|}{HOPE (Description-Based)} & \multicolumn{3}{c}{POPE (Adversarial Sampling)} \\
    \midrule
    \multicolumn{1}{c|}{Model} & Precision & Recall & F1 Score & Precision & Recall & F1 Score \\
    \midrule
    LLaVA Next 8B & 66.51  & 86.33  & 75.14  & 90.04  & 86.20  & 88.08  \\
    LLaVA OV 7B & 67.93  & 88.53  & 76.87  & 90.52  & 88.47  & 89.48  \\
    Qwen2-VL 7B & 76.00  & 81.93  & 78.86  & 93.11  & 82.00  & 87.20  \\
    Qwen2.5-VL 7B & 81.35  & 81.13  & 81.24  & 93.98  & 81.13  & 87.08  \\
    InternVL2.5 8B & 80.72  & 89.87  & 85.05  & 89.76  & 90.00  & 89.88  \\
    InternVL3 8B & 75.35  & 93.13  & 83.30  & 88.08  & 93.07  & 90.50  \\
    \bottomrule
    \end{tabular}%
  \label{tb:attr_coco}%
\end{table}%

\begin{table}[!t]
  \centering
  \caption{Results of the state-of-the-art LVLMs under description-based hallucination searching of HOPE on VG and OpenImage.}
    \begin{tabular}{l|ccc|ccc}
    \toprule
    \multicolumn{1}{c|}{Dataset} & \multicolumn{3}{c|}{VG} & \multicolumn{3}{c}{OpenImages} \\
    \midrule
    \multicolumn{1}{c|}{Model} & Precision & Recall & F1 Score & Precision & Recall & F1 Score \\
    \midrule
    LLaVA Next 8B & 66.40  & 83.13  & 73.83  & 66.19  & 80.13  & 72.50  \\
    LLaVA OV 7B & 67.52  & 84.80  & 75.18  & 69.97  & 84.20  & 76.43  \\
    Qwen2-VL 7B & 78.05  & 80.13  & 79.08  & 69.81  & 72.47  & 71.12  \\
    Qwen2.5-VL 7B & 82.13  & 84.87  & 83.48  & 83.22  & 78.33  & 80.70  \\
    InternVL2.5 8B & 75.33  & 84.47  & 79.64  & 80.77  & 80.93  & 80.85  \\
    InternVL3 8B & 77.31  & 87.00  & 81.87  & 82.07  & 80.27  & 81.16  \\
    \bottomrule
    \end{tabular}%
  \label{tb:attr_vg_oi}%
\end{table}%

Table \ref{tb:attr_coco} reports the results of different LVLMs under the description-based hallucination searching on MS-COCO. To validate the effectiveness of our proposed benchmark, we also report the performance of the same models under the adversarial setting of the POPE benchmark. As shown in the table, the distractors generated by the description-based searching are more effective in inducing hallucinations than those sampled using the adversarial strategy in POPE. Across different models, the precision drops by at least 9.2\% (on InternVL2.5) and up to 23.7\% (on LLaVA-OV), with an average decrease of 16.3\%. These results convincingly demonstrate that the description-based searching poses a more challenging evaluation scenario for LVLMs and serves as an effective strategy for assessing object hallucination. Table \ref{tb:attr_vg_oi} reports the results of different LVLMs under the description-based hallucination searching on VG and OpenImage. As shown in the table, the LLaVA series models exhibit weak robustness to the description-based searching, resulting in a large number of hallucinations and consistently low precision. In contrast, the Qwen-VL and InternVL series demonstrate relatively better performance, though there remains considerable room for improvement. Notably, InternVL3, despite being the most recently released model, achieves only 77.31\% and 83.24\% in precision on the VG and OpenImages datasets, respectively.

\subsection{Study on The Size of Distractor Space}

\begin{figure}[!t]
    \centering
    \subfigure{\includegraphics[width=0.45\textwidth]{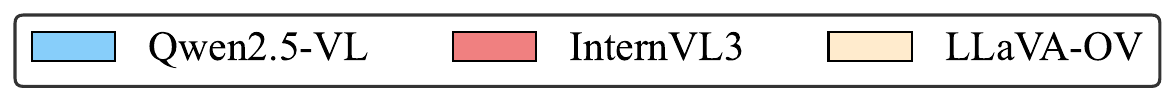}}
    \vspace{-1em}
    
    \setcounter{subfigure}{0}
    \subfigure[Co-occurrence]{\includegraphics[width=0.32\textwidth]{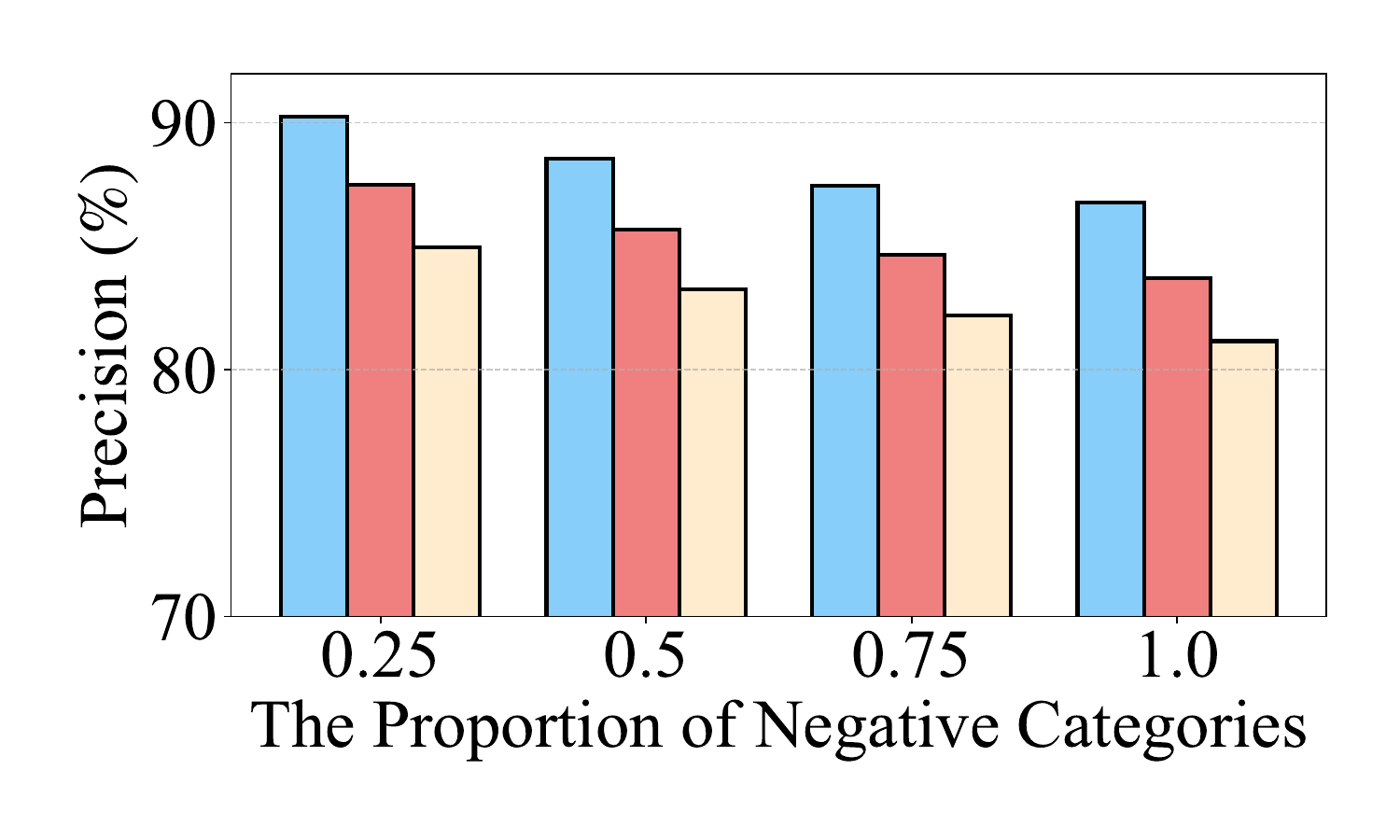}}
    \subfigure[Similarity]{\includegraphics[width=0.32\textwidth]{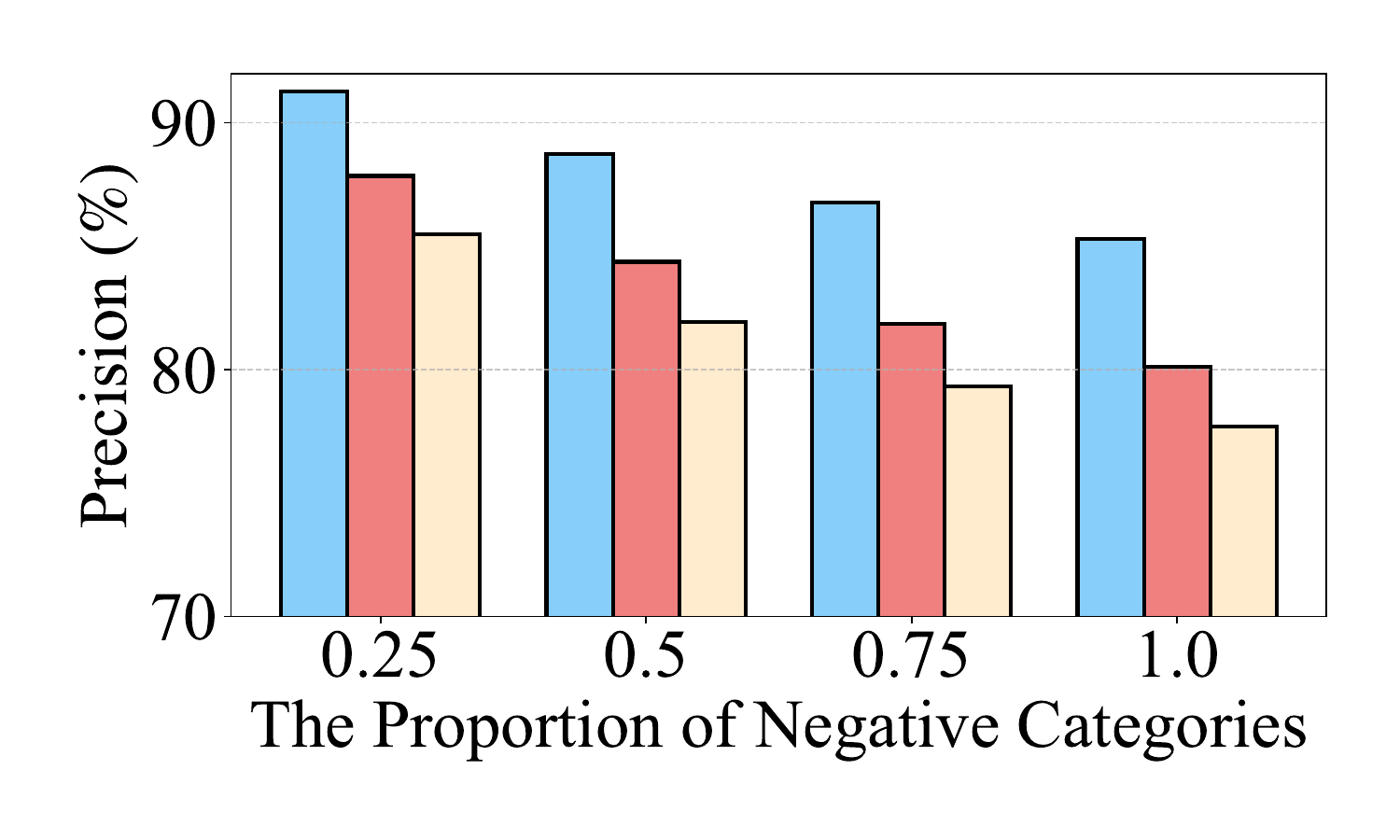}}
    \subfigure[Content-Aware]{\includegraphics[width=0.32\textwidth]{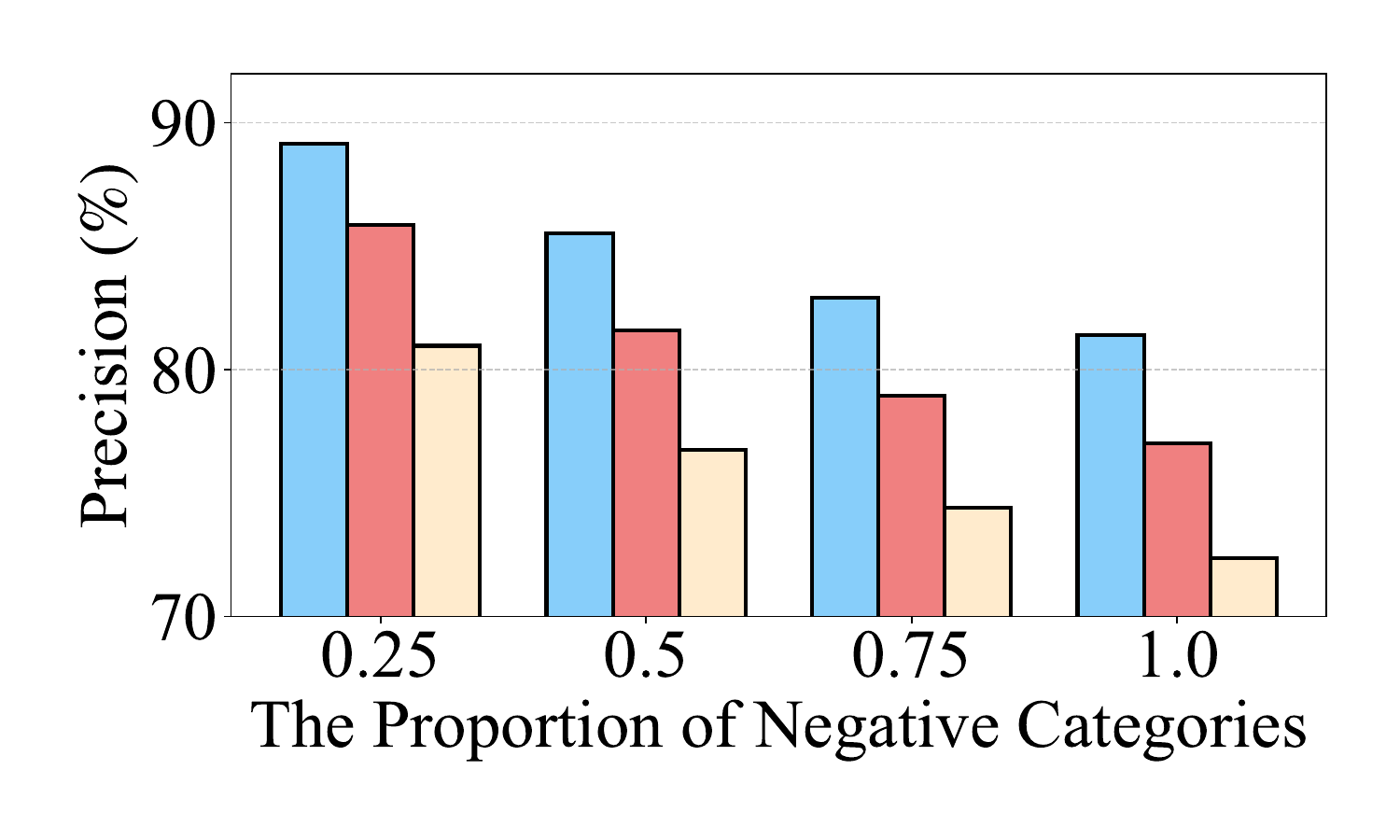}}
    \caption{Precision of different LVLMs under three types of hallucination searching as the size of the distractor space increases.}
    \label{fig:size}
\end{figure}

Next, we investigate the impact of the distractor space size on object hallucination evaluation. Our analysis focuses on the category-oriented and content-aware searching. Specifically, for each sample, we retain a certain proportion $\gamma$ of the negative categories, where $\gamma\in\{0.25,0.5,0.75,1.0\}$. Figure \ref{fig:size} presents bar charts showing the precision of different LVLMs as the value of $\gamma$ increases. As observed, models become more susceptible to hallucination as the distractor space expands. This finding highlights that the size of the distractor space plays a critical role in the effective evaluation of object hallucination. A larger distractor space is more likely to contain highly misleading distractors, making it easier to trigger hallucinations in LVLMs. Conversely, a smaller distractor space may fail to capture such challenging distractors. This observation also explains our motivation for constructing the benchmark on the Objects365 dataset, which contains a significantly larger number of categories than MS-COCO.

\begin{figure*}[!t]
    \centering
    {\includegraphics[width=0.95\textwidth]{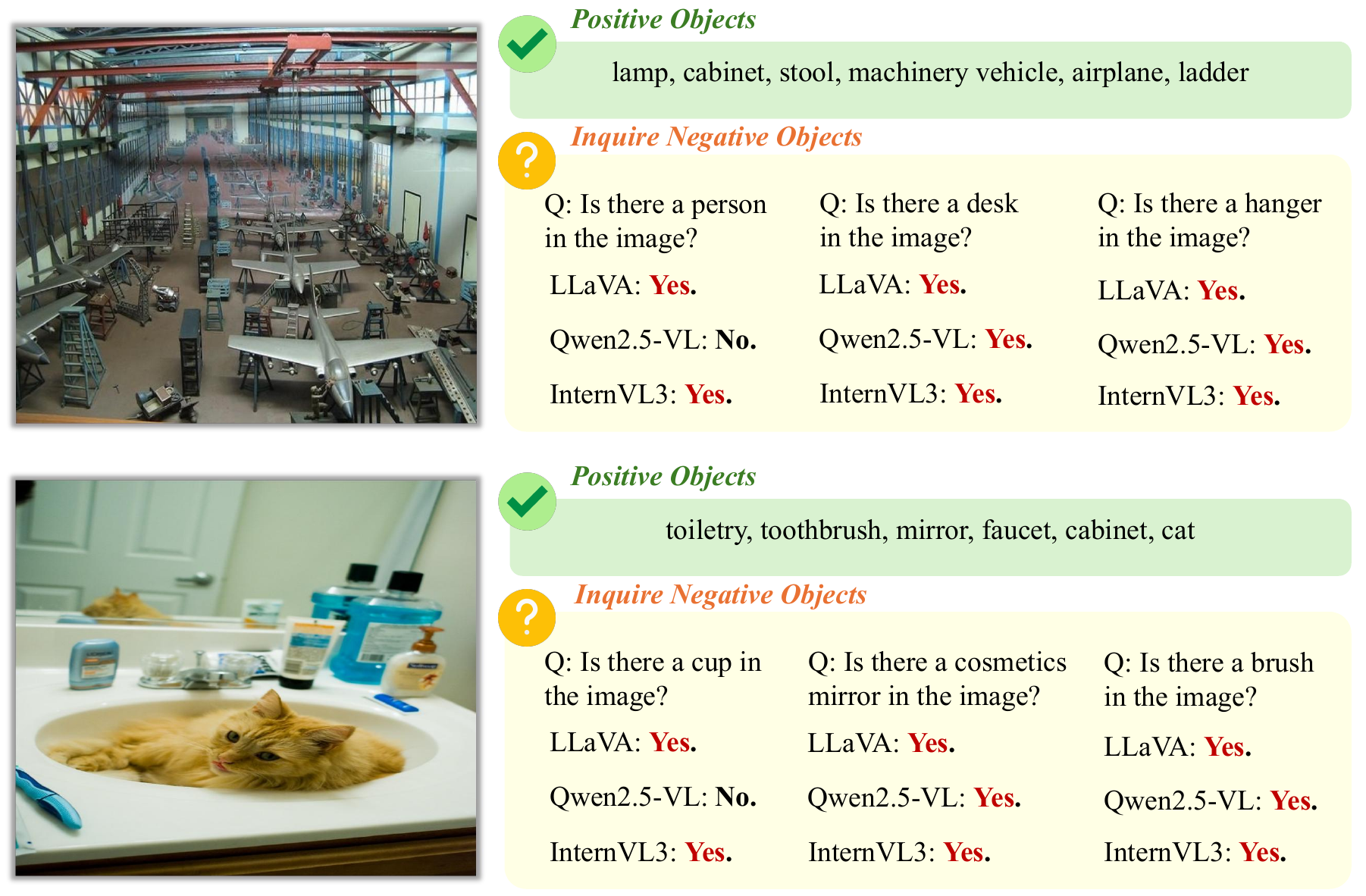}}
    \caption{Examples of three different searching strategies: co-occurrence, similarity and content-aware.}
    \label{fig:case}
\end{figure*}

\subsection{Visulization of Hallucination Searching}

To better understand the underlying mechanisms behind hallucination searching, Figure \ref{fig:case} presents illustrative examples under three different searching strategies: category-oriented (co-occurrence and similarity) and content-aware, shown from left to right. As observed in the figure, the co-occurrence-based searching targets the semantic co-occurrence patterns between object categories. It exploits prior knowledge about which objects are likely to appear together in a given image. In contrast, the similarity-based searching focuses on visual resemblance between object categories, inducing hallucination by introducing distractors that are visually similar to the ground-truth objects. Notably, both co-occurrence and similarity-based searching often operate independently of the actual visual content of the image. For instance, the inclusion of “person” in the first image and “cosmetics mirror” in the second are driven by such priors rather than the image itself. The content-aware searching, on the other hand, is explicitly grounded in the visual content. It identifies ambiguous or visually complex regions in the image and selects distractor categories that are most likely to be confused with the actual content. This makes it more directly aligned with the image-specific features and thus potentially more effective in inducing hallucination.

\begin{figure*}[!t]
    \centering
    {\includegraphics[width=0.99\textwidth]{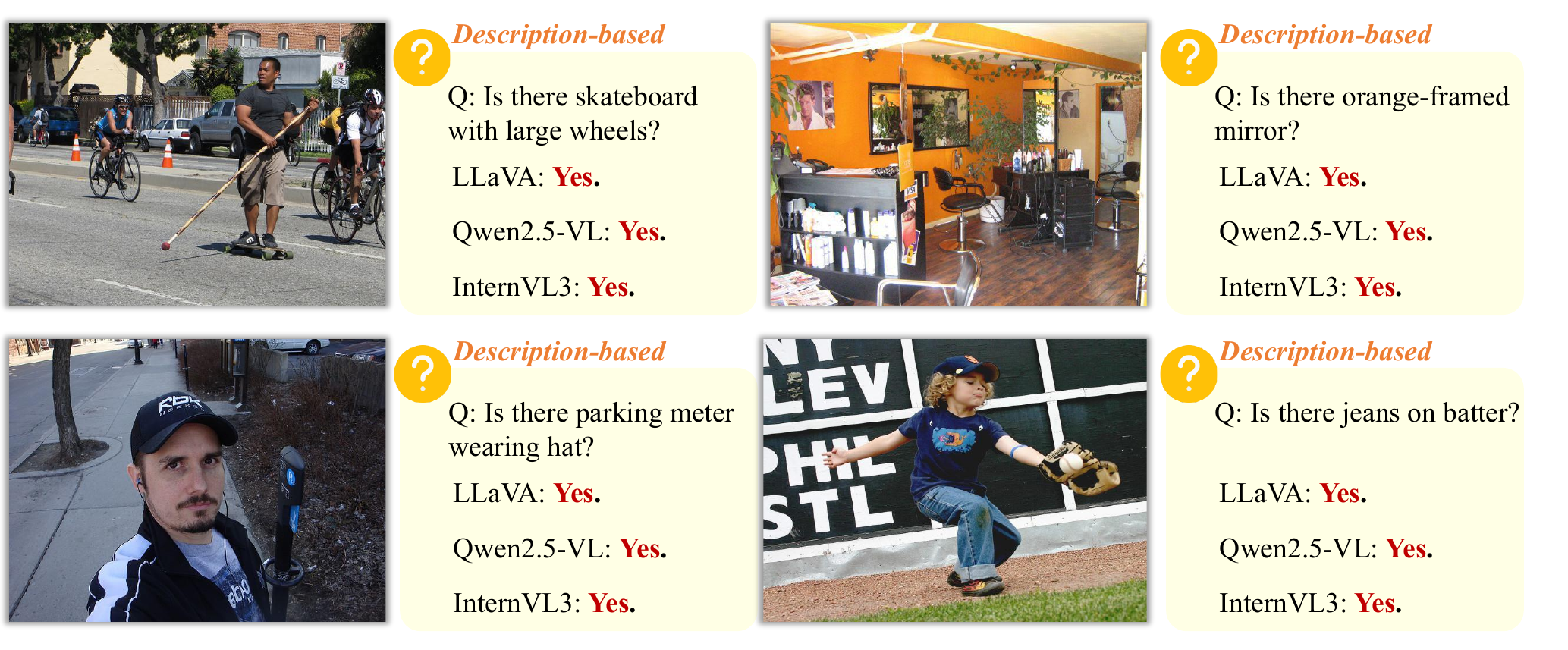}}
    \caption{Examples of the description-based hallucination searching.}
    \label{fig:attr_case}
\end{figure*}

Figure \ref{fig:attr_case} presents several examples of the description-based hallucination searching on the MS-COCO (first column) and VG (second column) datasets. Overall, although current LVLMs demonstrate strong object recognition capabilities, they remain susceptible to hallucinations when confronted with deceptive relational cues. For example, in the top-left image, all LVLMs confuse the skateboard with the bicycle wheel, failing to distinguish between the two. In the bottom-left image, none of the LVLMs correctly identify the relationship between the cap and the parking meter, leading to consistent hallucinations. These observations suggest that, compared to the previous two searching strategies, the description-based strategy poses a more challenging evaluation setting. By introducing misleading combinations, such as incorrect relationships, it more effectively exposes the vulnerabilities of LVLMs and serves as a powerful tool for hallucination assessment.

\section{Related Work}

With the growing interest in hallucination in LVLMs, a number of benchmarks have been proposed to evaluate this phenomenon. Among them, CHAIR \citep{rohrbach2018object} and POPE \citep{li2023evaluating} represent two of the most influential early efforts in the study of object hallucination. CHAIR evaluates the extent of object hallucination by measuring the proportion of hallucinated objects in generated image descriptions, which often requires complex human-crafted parsing rules to perform exact object matching. POPE adopted simple yes/no binary questions for evaluation and revealed that LVLMs tend to hallucinate objects that are frequent or co-occurring in their training data. NOPE, proposed by Lovenia \citep{lovenia2023negative}, used a predefined set of negative pronouns to detect hallucinations in model outputs. R-Bench \citep{wu2024evaluating} and H-POPE \citep{pham2024h} are two extensions of POPE that respectively evaluates attribute and relationship hallucinations. ROPE \citep{chen2024multi} focus on multi-object hallucination, revealing that LVLMs suffer more hallucinations when focusing on multiple objects compared to a single object.

\section{Conclusion}

In this paper, we introduced the HOPE benchmark for object hallucination evaluation in LVLMs. Specifically, we discussed three hallucination searching strategies: category-oriented, content-aware, and description-based. The category-oriented hallucination searching focuses on the relationships between object categories, leveraging two complementary cues: object co-occurrence (as adopted in POPE) and visual similarity. The content-aware hallucination searching incorporates image-specific information by identifying ambiguous regions within an image and selecting misleading negative object categories as distractors. Finally, the description-based hallucination searching expands the traditional distractor space by constructing combinations of true objects with false descriptions (\textit{e.g.}, incorrect attributes or states), and identifying those combinations that are most likely to induce hallucination. Experimental results with multiple state-of-the-art LVLMs validated the effectiveness of the proposed benchmark. In future, we plan to further expand the distractor space by incorporating elements such as object-relevant captions, aiming to enhance the misleading capacity of the generated distractors.

\bibliography{references}
\bibliographystyle{iclr2025_conference}

\newpage
\appendix

\begin{table}[]
    \centering
    \small
    \caption{Statistics on \textit{Description-Based Hallucination Searching} data.}
    \begin{tabular}{l|ccc|ccc}
    \toprule
    \multirow{2}{*}{\#Dataset} & \multicolumn{3}{c}{\#Filter Condition}  & \multicolumn{3}{|c}{\#Statistical Result} \\
                               & Object Frequency & Descriptions (/object) & Objects (/image) & Objects    & Descriptions & Samples    \\
    \midrule
    MS-COCO                    & -                   & -                 & -              & 80           & 1 - 1,884       &  40,137   \\
    VG                         & $\geq$ 2,000         & $\geq$ 50         & $\geq$ 10      & 265          & 90 - 4,401   &  88,730   \\
    OpenImages                 & $\geq$ 1,000         & $\geq$ 50         & $\geq$ 10      & 97           & 71 - 2,009   &  4,435  \\
    \bottomrule
    \end{tabular}
    \label{tab:descriptions}
\end{table}

\section{Implementation Details}
Regarding the construction of the HOPE dataset, we developed a simple and extensible toolkit featuring an easy-to-read code structure and a rich set of hyperparameters, allowing users to reproduce or extend existing question-answering (QA) data. Users can independently choose data sources, sampling quantities, searching strategies, and prompt templates for constructing QA pairs. The complete toolkit is included in the supplementary materials.

When constructing \textit{Category-Oriented Hallucination Searching} and \textit{Content-Aware Hallucination Searching} data, unlike POPE—which sequentially selects one positive object at a time along with its most relevant negative object under a given searching strategy—HOPE randomly selects a specified number of positive objects from a given sample at one time, ranks all negative objects based on the scores of the specified searching strategy, and then selects the top-\textit{k} negatives. This approach effectively avoids situations where certain categories lack suitable corresponding negative objects under some searching strategies. For instance, in the category space of the COCO dataset, the class ``person'' struggles to find its most similar counterpart when using a similarity-based searching strategy. Moreover, HOPE offers greater flexibility in selecting negative objects, unlike POPE which strictly requires the number of negative objects to match that of positive objects.

For constructing \textit{Description-Based Hallucination Searching} data, we follow a systematic pipeline: 

\begin{enumerate}[label=(\roman*)]
    \item Data Cleaning: First, we standardized category names in the LSA dataset, merged multiple identical objects within the same image, and filtered out objects with too few descriptions or low occurrence frequencies in the LSA dataset, while ensuring each image contains a sufficient number of objects. The filtering criteria and results are presented in Table \ref{tab:descriptions}. Taking the VG dataset as an example, the table shows that we required category objects in the LSA dataset's VG data to appear in at least 2,000 images, with each object containing at least 50 descriptions, and each image needed to contain at least 10 objects meeting these conditions. Ultimately, we obtained qualified VG data comprising 265 categories, with description counts ranging from 90 to 4,401, and 88,730 qualifying images. Specifically, to ensure a fair comparison with POPE experiments, we fixed the number of COCO categories in the LSA dataset to 80, consistent with POPE's settings. 
    \item Strategy Application: From the filtered data, we selected 500 homologous samples as benchmark examples. For each image, we paired positive objects with negative descriptions, input them into the CLIP model to calculate scores, and typically selected the highest-scoring pair as the most misleading distractor. 
    \item Human Verification: For all combinations ranked by score from the previous step, we performed manual validation to modify descriptions that, while having the highest scores, were illogical (such as combinations containing grammatical errors). 
    \item QA Generation: After manual checking, we inserted the verified most misleading distractors (positive object-negative description pairs) into given prompt templates to generate QA data.
\end{enumerate}

We employed the fixed random seed (=1) to ensure experimental reproducibility, with more detailed parameters and implementation code available in the supplementary materials.

\end{document}